\pdfoutput=1

\documentclass[11pt]{article}

\usepackage{emnlp2021}

\usepackage{times}
\usepackage{latexsym}

\usepackage[T1]{fontenc}

\usepackage[utf8]{inputenc}

\usepackage{microtype}

\usepackage{amsmath}
\usepackage{amssymb}
\usepackage{url}
\usepackage{booktabs} 
\usepackage{graphicx}
\usepackage{epstopdf}
\usepackage{subfigure}
\usepackage{verbatim}
\usepackage{bm}
\usepackage{array}
\usepackage{multirow}
\usepackage{subfigure}
\usepackage{makecell}
\usepackage{xcolor}
\usepackage{xspace}
\usepackage[most]{tcolorbox}

\usepackage[font={small}]{caption}

\newcommand{\dataset}[0]{\textsc{FinQA}\xspace}

\title{\textsc{FinQA}: A Dataset of Numerical Reasoning over Financial Data}

\author{\textbf{Zhiyu Chen}\textsuperscript{1}, \textbf{Wenhu Chen}\textsuperscript{1}, \textbf{Charese Smiley}\textsuperscript{2}, \textbf{Sameena Shah}\textsuperscript{2}, \\ \textbf{Iana Borova}\textsuperscript{1}, \textbf{Dylan Langdon}\textsuperscript{1}, \textbf{Reema Moussa}\textsuperscript{1}, \textbf{Matt Beane}\textsuperscript{1}, \textbf{Ting-Hao Huang}\textsuperscript{3}, \\  \textbf{Bryan Routledge}\textsuperscript{4} and \textbf{William Yang Wang}\textsuperscript{1}\\
  \textsuperscript{1}University of California, Santa Barbara \\
  \textsuperscript{2}J.P. Morgan \\
  \textsuperscript{3}Pennsylvania State University \\
  \textsuperscript{4}Carnegie Mellon University \\
  {\tt \{zhiyuchen,william\}@cs.ucsb.edu}}

\begin{document}
\maketitle

\begin{abstract}
The sheer volume of financial statements makes it difficult for humans to access and analyze a business's financials.
Robust numerical reasoning likewise faces unique challenges in this domain. 
In this work, we focus on answering deep questions over financial data, aiming to automate the analysis of a large corpus of financial documents.
In contrast to existing tasks on general domain, the finance domain includes complex numerical reasoning and understanding of heterogeneous representations.
To facilitate analytical progress, we propose a new large-scale dataset, \textbf{\textsc{FinQA}}, with \underline{\textbf{Q}}uestion-\underline{\textbf{A}}nswering pairs over \underline{\textbf{Fin}}ancial reports, written by financial experts.
We also annotate the gold reasoning programs to ensure full explainability.
We further introduce baselines and conduct comprehensive experiments in our dataset. 
The results demonstrate that popular, large, pre-trained models fall far short of expert humans in acquiring finance knowledge and in complex multi-step numerical reasoning on that knowledge.
Our dataset --- the first of its kind --- should therefore enable significant, new community research into complex application domains. The dataset and code are publicly available\footnote{\url{https://github.com/czyssrs/FinQA}}.
\end{abstract}

\section{Introduction}
\begin{figure*}[ht]
\centering
\includegraphics[width=\textwidth]{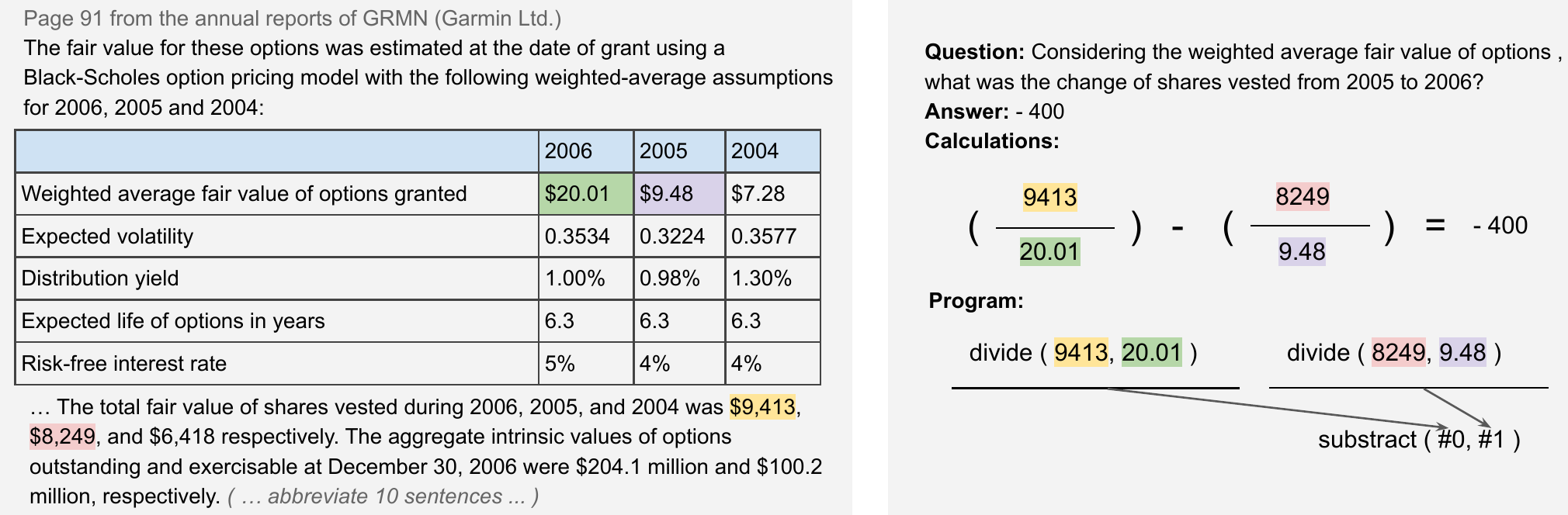}
\caption{An example from \textsc{FinQA}: The system needs to learn how to calculate the number of shares, then select relevant numbers from both the table and the text to generate the reasoning program to get the answer. } 
\label{fig:eg-intro}
\end{figure*}

Financial analysis is a critical means of assessing business performance, and the consequences of poor analysis can involve costs of billions of dollars~\cite{jerven2013poor, mackenzie2008engine}.
To facilitate high quality, timely decision making, professionals --- such as analysts or investors --- perform complex quantitative analysis to select information from financial reports.
Such analysis demands advanced expertise in reasoning among heterogeneous (structured and unstructured) data sources and performing complex numerical reasoning, for example, comparing financial ratios of profitability or growth.
These challenges are compounded by an exponentially expanding collection of company financial documents~\cite{mackenzie2012drilling,lange2016cultures} such that it is genuinely unclear whether dedicated human effort can produce fiscal analysis of sufficient quality for current decision making. 
This poses an interesting question: can we automate such deep analysis of financial data?

A few NLP studies in Question Answering (QA) explored the numerical reasoning capabilities needed to answer questions correctly.
For example, the DROP dataset~\cite{DBLP:conf/naacl/DuaWDSS019} focused on Wikipedia-based questions that require numerical reasoning, {\em e.g.,} ``Where did Charles travel to first, Castile or Barcelona?'' needs a comparison between the times of two events. 
However, most prior work only targeted the general domain,
where the questions involve much less calculation (mostly one-step calculation) than that of the financial domain.
Financial QA is more challenging than classic QA~\cite{DBLP:conf/acl/RajpurkarJL18, DBLP:conf/emnlp/Yang0ZBCSM18} because it requires the system to spot relevant information across heterogeneous sources, such as tables and unstructured texts, and then create a numerical reasoning path to connect all the information.
It also takes substantial knowledge to ask meaningful financial questions.
It is not clear how well the large language models, which performed well for general-domain QA, can be adapted to answer realistic, complex financial questions.

This paper introduces \textbf{\textsc{FinQA}}, a \textbf{expert-annotated} dataset that contains 8,281 financial QA pairs, along with their numerical reasoning processes.
Eleven finance professionals collectively constructed \dataset based on the earnings reports of S\&P 500 companies~\cite{fintabnet}.
The questions in \dataset, such as ``Considering the weighted average fair value of options, what was the change of shares vested from 2005 to 2006?'' (Figure~\ref{fig:eg-intro})
and 
``What was the net change in tax positions in 2014?'',
require information from both tables and unstructured texts to answer.
The reasoning processes answering these questions are made of many common calculations in financial analysis, such as addition, comparison, and table aggregation.
To the best of our knowledge, \dataset is the first dataset of its kind to tackle complicated QA tasks based on the real-world financial documents.

We propose a retriever-generator QA framework to first retrieve supporting facts from financial reports, then to generate executable reasoning programs to answer the questions.
Equipped with pre-trained language models, such as BERT~\cite{DBLP:conf/naacl/DevlinCLT19} and RoBERTa~\cite{DBLP:journals/corr/abs-1907-11692}, our proposed approach outperforms all other baselines and achieves an execution accuracy of 65.05\%.
Although our system outperforms the non-expert crowd (50.68\%), the significant accuracy gap between the model and human experts (91.16\%) motivates the need for future research.

The main contribution of this work is three-fold:


\begin{itemize}
    \item We propose the task of QA over financial data to assist financial analysis. The task emphasizes an important phenomenon for the NLP community to study and analyze how the current pre-trained models perform on complex and specialized domains. 
    \item We construct a new large-scale dataset, \textsc{FinQA}, with 8,281 examples written by financial experts, with fully annotated numerical reasoning programs. 
    \item We experiment on various baselines and find that the models are still far behind expert performance, strongly motivating future research.
\end{itemize}

\section{Related Work}

\paragraph{Questions Answering.} There have been several QA datasets involving numerical understandings and calculations. The major source is from structured tables or knowledge bases, owning the nature to succinctly organize numerical information. Popular datasets include ComplexWebQuestions~\cite{DBLP:conf/naacl/TalmorB18}, WikiTableQuestions~\cite{DBLP:conf/acl/PasupatL15}, Spider~\cite{DBLP:conf/emnlp/YuZYYWLMLYRZR18}, TabFact~\cite{DBLP:conf/iclr/ChenWCZWLZW20}, etc. For reading comprehension, the dataset most related to ours is the DROP dataset~\cite{DBLP:conf/naacl/DuaWDSS019}, which applies simple calculations over texts. The top methods on DROP typically use specific prediction heads for each kind of calculation. HybridQA~\cite{DBLP:conf/emnlp/ChenZCXWW20} targets QA over both the table and the text, but not with the focus of numerical reasoning. All these existing datasets are built upon the general domain (mostly based on Wikipedia). In contrast, our dataset focus on the finance domain, which demonstrates much more complex nature in numerical reasoning questions, combining both the structured tables and unstructured texts. Another kind of QA datasets related to ours is the math word problem datasets, like MaWPS~\cite{DBLP:conf/naacl/Koncel-Kedziorski16}, MathQA~\cite{DBLP:conf/naacl/AminiGLKCH19}. The task is to generate the solution programs given a short input math problem. Existing models include~\cite{DBLP:conf/emnlp/KimKLG20, DBLP:conf/icml/ChenHPSFG20, DBLP:conf/iclr/ChenLYZSL20}, etc. 


\paragraph{Financial NLP.} Financial NLP has become one of the major application domains attracting growing attentions. Previous works in finance domain include risk management to detect fraud~\cite{DBLP:conf/acl/HanBHDBW18, DBLP:conf/emnlp/WangZLZL19, DBLP:journals/corr/abs-1908-09156}, sentiment analysis to assist market prediction~\cite{DBLP:conf/asunam/DayL16,DBLP:conf/ijcnlp/WangTLC13,DBLP:conf/emnlp/AkhtarKGEB17}, 
opinionated Question Answering~\cite{DBLP:conf/ijcai/0001HH0Z20}, such as the FiQA\footnote{https://sites.google.com/view/fiqa/home} dataset built from forums and social media. Recent works attempt to develop pre-trained models specialized for finance domain~\cite{DBLP:journals/corr/abs-2006-08097, DBLP:journals/corr/abs-1908-10063}, and the downstream tasks are mostly sentiment classifications. 
To the best of our knowledge, there is no previous work and dataset on building QA systems of numerical reasoning on financial reports. 

\section{Task Definition}
\label{task_def}

\paragraph{Problem Formulation.}
Presented with a financial report consisting of textual contents $E$ and structured table $T$, given a question $Q$, the task is to generate the reasoning program $G=\{w_0, w_1, ... w_n\}$, where ${w_i}$ is the program tokens defined by domain specific language (DSL), then it is executed to get the answer A:
\begin{equation}
    P(A | T, E, Q) = \sum P(G_i | T, E, Q)
\end{equation}
Where $\{G_i\}$ is all the correct programs to evaluate to the answer. For financial tables, there is typically a description header (blue header in Figure~\ref{fig:eg-intro}), which often gives the timing information; and each row has its name on the left. Some of the financial tables may demonstrate more complicated layouts, {\em e.g.}, nested structures. 
As a first step for this direction, in this paper we only focus on the regular layout cases for simplicity.

\paragraph{Domain Specific Language.}
In this work, we use DSL consisting of mathematical operations and table operations as executable programs. 
The program consists of a sequence of operations:
\begin{equation}
    \text{op}_{1} [\boldsymbol{\text{args}_{1}}] , \text{op}_2 [\boldsymbol{\text{args}_{2}}] ... , \text{op}_n [\boldsymbol{\text{args}_{n}}]
\end{equation}

Each operation takes a list of arguments $\boldsymbol{args_n}$. On consulting with financial experts, as most of the accounting and financial valuation theory primarily include linear algebra, we include 10 common types of operations in our dataset. There are 6 mathematical operations: \texttt{add}, \texttt{subtract}, \texttt{multiply}, \texttt{divide}, \texttt{greater}, \texttt{exp}, 
and 4 table aggregation operations \texttt{table-max}, \texttt{table-min}, \texttt{table-sum}, \texttt{table-average}, 
that apply aggregation operations on table rows. The mathematical operations take arguments of either numbers from the given reports, or a numerical result from a previous step; The table operations take arguments of table 
row names. We use the special token $\#n$ to denote the result from the $n$th step. 
For example, in Figure~\ref{fig:eg-intro}, the program consists of 3 steps; The first and the second division steps take arguments from the table and the text, respectively, then the third step subtracts the results from the two previous steps. 
Refer to Appendix A for more details of the operations and the grammars. 

\paragraph{Evaluations.}
Previous studies on QA with numerical reasoning only evaluate the execution accuracy, i.e., the final results from the generated programs, such as DROP~\cite{DBLP:conf/naacl/DuaWDSS019} and MathQA~\cite{DBLP:conf/naacl/AminiGLKCH19}. However, the applications for the  finance domain generally pose much higher requirements of explainability and transparency. Therefore, we also provide the gold programs for our dataset. Besides execution accuracy, we also propose to evaluate the accuracy of the generated programs. Specifically, we replace all the arguments in a program with symbols, and then we evaluate if two symbolic programs are \textit{mathematically equivalent}. For example, the following two programs are equivalent programs:
\begin{align*}
\begin{array}{l}
    \text{add}(a_1, a_2), \text{add}(a_3, a_4), \text{subtract}(\#0, \#1) \\
    \text{add}(a_4, a_3), \text{add}(a_1, a_2), \text{subtract}(\#1, \#0)
\end{array}
\end{align*}
Note that execution accuracy tends to overestimate the performance because sometimes the model just hit the correct answer by chance; While program accuracy tends to produce false negatives since some questions may have multiple correct programs.
\section{The \textsc{FinQA} Dataset}

\subsection{Data Preparation}
\label{subsec:data-preprocessing}

\paragraph{Data Source.}
We develop \textsc{FinQA} based on the publicly available earnings reports of S\&P 500 companies from 1999 to 2019, collected in the FinTabNet dataset~\cite{fintabnet}.
An earnings report is a set of pages in a PDF file that outlines the financials of a company, which usually contains tables and texts.
The FinTabNet dataset has annotated the tables in each report.

\paragraph{Data Filtering.}
Realistic earnings reports contain many tables not suitable for numerical reasoning tasks. 
Equipped with the table annotations in FinTabNet, we filter the data as follows:
First, we extract the pages in earnings reports with at most one table.
Second, we exclude the tables with over 20 rows, over 2 description headers, or with other complex nested structures. 
We also exclude the tables with tedious contents, such as catalogs, which is common in FinTabNet. 
As stated in \S\ref{task_def}, these over-complicated tables are out of the scope of this work.
Finally, for the tables with 2 description headers, we merge them into a single header to simplify the representations.
As a result, a total of 12,719 pages were selected for further annotation.

\subsection{Annotation Procedure}
\label{subsec:annotation-procedure}

\paragraph{Recruiting Expert Annotators.}
We post job ads on UpWork\footnote{UpWork (www.upwork.com) is a platform where requesters can recruit skilled freelancers.} and hire eleven US-based experts with professional finance backgrounds (CPAs, MBAs, etc.)
Each hire is interviewed using four example report pages and asked to compose example Q\&A pairs.
After hiring, each annotator first goes through a training session to learn the task and the annotation interface (Appendix D).
When the workers fully master the annotation process, we launch the official batches for them to work on.

An annotator can compose up to two questions for each given report page or skip if it is hard to compose any meaningful question.
We pay around \$2.0 for each question, which leads to an average hourly wage of \$35.0. 
The whole data collection took around eight weeks. 

We do not use popular micro-task platforms, such as Amazon Mechanical Turk (MTurk), because our preliminary studies show that many MTurk workers can not perform this task effectively.
Our experiment with MTurk workers in \S~\ref{sec:label-quality-eval} further echo this observation.
As most existing QA datasets were constructed by MTurk workers~\cite{DBLP:conf/emnlp/Yang0ZBCSM18, DBLP:conf/naacl/DuaWDSS019, DBLP:conf/emnlp/ChenZCXWW20}, it requires substantial domain-specific knowledge to compose meaningful questions that are hard for computers to answer.


\paragraph{Annotation Task Design.}
For each page selected in \S\ref{subsec:data-preprocessing}, the annotators are asked to 
{\em (i)} write a meaningful financial question, 
{\em (ii)} compose a reasoning program to answer the question, and
{\em (iii)} to annotate the supporting fact.
Each page is assigned to one or two experts for annotation. 
We detail each part as follows.
\textbf{(I) Financial question:}
For a given page of earnings reports, the annotators are asked first to compose a question that is ``meaningful for financial analysis or learning insights of the company financial reports'' and require numerical calculations to answer.
We encourage the experts to write questions that require the information from both the text and the table to answer.
\textbf{(II) Reasoning program:}
After providing the question, the annotators are then asked to elaborate the operation steps to answer the question.
Specifically, they compose a maximum of 5 steps of operation, where each operation has four slots: ``operation'', ``argument1'', ``argument2'', and ``result''.
The ``operation'' is one of the ten predefined operations described in \S\ref{task_def}.
An ``argument'' is a number or a table's row name, either from the report or a previous step's result.
For operations that only use one argument, such as table aggregation, workers can leave argument2 blank. 
The annotation interface (see Appendix D) automatically validates the inputs to ensure correctness.
\textbf{(III) Supporting fact:} 
We also ask the annotators to mark all the sentences in the text and the table rows that contain the information needed to answer the question.

\subsection{Data Quality Assessment}
\label{sec:label-quality-eval}

\paragraph{External experts answer \textsc{FinQA} questions with a high accuracy and a high inter-annotator agreement.}
To validate the quality of the annotations, as well as to set up human expert performance upper bound, we hire another two financial professionals on UpWork.
We randomly sample 200 examples from our dataset, and ask the professionals to answer the questions as well as write the operation steps, following the same procedure as in the dataset construction. The payment is \$2.0 per question. 
For execution accuracy, they reach 92.25\% and 90.06\%, respectively (mean = 91.16\%).
For program accuracy, they reach 89.44\% and 85.53\% (mean =  87.49\%).
The agreements between the two annotators are 92.65\% for execution accuracy, and 86.76\% for program accuracy. 

\paragraph{Non-expert crowd workers answer \textsc{FinQA} questions with a low accuracy.}
We also test how well non-expert MTurk workers can answer \textsc{FinQA} questions.
We distribute the samples to MTurk\footnote{Three built-in worker qualifications are used: HIT Approval Rate ($\geq $95\%), Number of Approved HITs ($\geq 3000$), and Locale (US Only) Qualification. We do not select any profession constraints. We pay \$2.0 for each question.}
and take the similar process to distribute each example to two workers.
We end up with an average execution accuracy of 50.68\% and a program accuracy of 48.17\%, which is far below the expert performance; the agreement rate is only around 60\%. 
These results echo our preliminary study's observations for MTurk workers in \S\ref{subsec:annotation-procedure}.

\subsection{Data Analysis}
\label{data_analysis}

\textsc{FinQA} contains 8,281 examples.
The data is released as training (6,251), validation (883), and test (1,147) following an 75\%/10\%/15\% split. 
The three sets do not have overlapping input reports. 
We quantitatively analyze some key properties of \textsc{FinQA}. 
Table~\ref{table:gen_stats} shows the general statistics.

\begin{table}[t]
\small
\begin{center}
\resizebox{0.45\textwidth}{!}{%
\begin{tabular}{lr}
\toprule
Examples (Q\&A pairs with program, fact) & 8,281\\
Report pages & 2,789\\
Vocabulary & 22.3k\\
Avg. \# sentences in input text & 24.32\\
Avg. \# tokens in input text & 628.11\\
Avg. \# rows in input table & 6.36\\
Avg. \# tokens in input table & 59.42\\
Avg. \# tokens in all inputs (text \& table) & 687.53\\
Max. \# tokens in all inputs (text \& table) & 2,679\\
Avg. question length & 16.63\\
\bottomrule
\end{tabular}
}
\caption{Statistics of \textsc{FinQA}.}
\label{table:gen_stats}
\end{center}
\end{table}

\paragraph{Statistics of Supporting Facts.}
In \dataset,
23.42\% of the questions only require the information in the text to answer; 
62.43\% of the questions only require the information in the table to answer;
and 14.15\% need both the text and table to answer.
Meanwhile,
46.30\% of the examples have one sentence or one table row as the fact; 
42.63\% has two pieces of facts; 
and 11.07\% has more than two pieces of facts.
For the examples with more than one piece of fact, we also calculate the maximum distances between all the same example's facts. 
55.48\% has a maximum distance of 3 or less sentences\footnote{For tables, we consider one row as one ``sentence''.};
24.35\% has a maximum distance of 4-6 sentences; 
and 20.17\% has over 6 sentences.

\paragraph{Statistics of Reasoning Programs.}
In the programs, the most frequent operations, \texttt{add}, \texttt{subtract}, \texttt{multiply}, and \texttt{divide}, have the distributions of 14.98\%, 28.20\%, 5.82\%, and 45.29\%, respectively.
The operation \texttt{division} has the highest frequency, as calculating ratios is common in financial analysis.
In \dataset,
59.10\% of the programs have 1 step, 32.71\% have 2 steps, and the rest 8.19\% have 3 or more steps. 

\section{Baseline Systems}
In this section, we first describe our main baseline framework \textbf{FinQANet} in \S\ref{main_method}, and then we introduce other baselines in \S\ref{other_method}. 
\begin{figure}[t]
\centering
\includegraphics[width=0.48\textwidth]{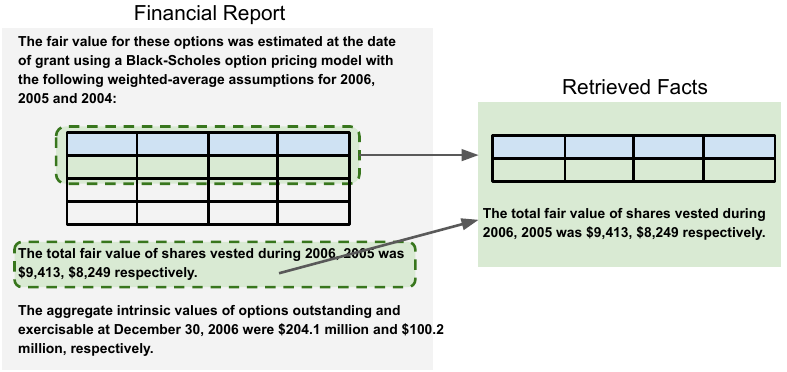}
\caption{The retriever retrieves supporting facts (text sentences or table rows) from the input financial report. } 
\label{fig:model-retriever}
\end{figure}
\subsection{The \textbf{FinQANet} Framework}
\label{main_method}
As a preliminary attempt on \textsc{FinQA}, we propose \textbf{FinQANet}, with a retriever to first retrieve the supporting facts from the input financial report, then a generator to generate the program to get the answer. 

\paragraph{Retriever} The full page of the financial report can go beyond 2,000 tokens, which cannot be coped with the current popular QA models~\cite{DBLP:conf/naacl/DevlinCLT19}. Therefore we first retrieve the supporting facts from the input report. For the tables, we use templates to turn each row into sentences. For example, the last row of the table in Figure~\ref{fig:eg-intro} is represented as `the risk-free interest rate of 2006 is 5\%; ...'. We concatenate each supporting fact with the question and train a classifier using pre-trained LMs like BERT~\cite{DBLP:conf/naacl/DevlinCLT19}. Then we take the top n retrieved facts, reordered as they appear in the input report. This set of retriever results will serve as the input to the second phase. Figure~\ref{fig:model-retriever} illustrates the retrieving procedure. 
Another common strategy is sliding window~\cite{DBLP:journals/corr/abs-1901-08634}. We take the sliding window of a fixed size with a stride to go through the report, then the windows containing all the supporting facts are marked as positive. However, we observe in the experiments that the length of the input to the program generator in the second phase greatly influences the performance. The performance of using sliding window falls far behind the previous method. 
\begin{figure*}[t]
\centering
\includegraphics[width=\textwidth]{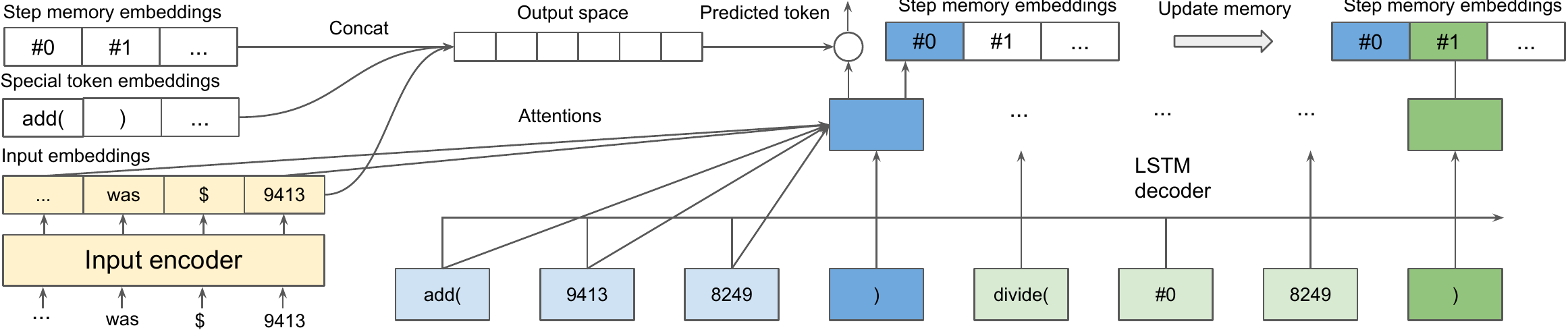}
\caption{The program generator. The retriever results and the question are first encoded using pre-trained LMs. At each decoding step, the model can generate from the numbers or table row names from the input, the special tokens in the DSL, or the step memory tokens. At the end of the generation of each operation step, we update the step memory token embeddings. } 
\label{fig:model-generator}
\end{figure*}
\paragraph{Program Generator}
Given the retrieved supporting facts from the retriever, the program generator aims to generate the executable program to answer the question. Figure~\ref{fig:model-generator} gives an overview of the program generator. The generated tokens come from 3 sources: 1) The input passage (retriever output) and the question tokens $\{e_i\}$, like the numbers or the table row names. 2) The special tokens $\{s_i\}$ from the DSL, like the function names, predefined constants, etc. 3) The step memory tokens $\{m_i\}$ to denote the results from previous steps, like $\#0$, $\#1$ , etc. We first use pre-trained LMs to encode $\{e_i\}$, denote the output embeddings as $\{h^e_i\}$. The embeddings of the special tokens and the step memory tokens are randomly initialized and denoted as $\{h^s_i\}$ and $\{h^m_i\}$ respectively. Denote all the token embeddings $H = [h^e_i; h^s_i; h^m_i]$. 

An LSTM is used for decoding. At each decoding step $T$, the program token embeddings $H$ are fed as the input; The decoder output $h_T$ is used to calculate the attention vector $att_p$ and $att_h$ over the input and the decoding history. Then a context vector $c_T$ combines all the contextual information:
\begin{equation}
    c_T = W_c [att_p; att_h; h_T]
\end{equation}
Meanwhile, another attention vector $att_p^{'}$ over the input is applied to all the token embeddings:
\begin{equation}
    H_T^{'} = W_h [H; H \circ att_p^{'}]
\end{equation}
Different from other program tokens, the step memory tokens $\{m_i\}$ imply the reasoning path of the program. To make use of such structure information, at each decoding step indicating the end of one $operation [args]$ unit, i.e., the step to generate the ending parentheses in our DSL, we compute another context vector $a_T$:
\begin{equation}
    a_T = W_a [att_p; att_h; h_T]
\end{equation}
Then the step memory token embedding corresponding to the current step is updated as $a_T$. 

The final prediction is calculated with:
\begin{equation}
    w_T = softmax(H_T^{'} \cdot c_T)
\end{equation}
During inference time, based on the grammar of the DSL, we use masks at each decoding step to ensure the structural correctness of the generated programs. In the retriever phase, we take the top n retrieved results as the input to the program generator. Therefore, for the training of the program generator, we use the retriever result on the training set (combined with the gold facts if there is any wrong prediction) as the input.

\subsection{Other Baselines}
\label{other_method}

\paragraph{TF-IDF + Single Op.}
We use TF-IDF to retrieve the top 2 sentences from the input report. Since the most common case in our dataset is one-step program and the most common operation is division, we take the first number from each sentence and apply the division operation. 

\paragraph{Retriever + Direct Generation.}
To demonstrate the necessity of generating the reasoning programs, we keep the architecture the same as our model, but directly generating the final results.

\paragraph{Retriever + Seq2seq.}
We use a Seq2seq architecture for the generator, similar to the Seq2seq baseline in the MathQA dataset~\cite{DBLP:conf/naacl/AminiGLKCH19}.
A bi-LSTM is used for encoding the input, and then an LSTM is used for decoding with attention. 

\paragraph{Retriever + NeRd.}
The Neural Symbolic Reader(NeRd)~\cite{DBLP:conf/iclr/ChenLYZSL20} is also a pointer-generator based model for program generation, with the state of the art results on the MathQA dataset~\cite{DBLP:conf/naacl/AminiGLKCH19}. Different from ours, it directly learns the program with nested format as a sequence, i.e., without the step memory tokens. This way the model is able to learn the program structures as patterns from very large-scale data (\textasciitilde40k for MathQA), but may fail on learning the reasoning paths. We keep the retriever part the same and compare with the generator part to demonstrate the usefulness of structure learning. 

\paragraph{Pre-Trained Longformer.}
There are also works on modeling very long documents with thousands of characters, with the attention mechanism that scales linearly with sequence
length, like the Longformer~\cite{DBLP:journals/corr/abs-2004-05150}. To demonstrate the necessity of breaking up into the pipeline of retriever and program generator, we remove the retriever and directly use the pre-trained Longformer as the input encoder in the program generator, and encode the whole report. The table rows are linearized similar as in \S\ref{main_method}. 
\section{Experimental Results}

\paragraph{Experiment Setups.}
For the retriever, we use BERT-base as the classifier (other pre-trained models perform similarly). Since most of the examples in our dataset have 1 or 2 facts, and we find that longer inputs lower the performance of the program generator, we take the top 3 ranked facts as the retriever results. For the program generator, we experiment on using BERT~\cite{DBLP:conf/naacl/DevlinCLT19}, RoBERTa~\cite{DBLP:journals/corr/abs-1907-11692}, and FinBert~\cite{DBLP:journals/corr/abs-1908-10063} as the encoder, to test the performances of popular large pre-trained models. For all models, we use the Adam optimizer~\cite{DBLP:journals/corr/KingmaB14}. Check Appendix B for more details of training and parameter settings. 

\subsection{QA Model Performance}

Table~\ref{table:main_res} presents the results for all the baseline systems. We evaluate the execution accuracy (exe acc) and program accuracy (prog acc) as explained in \S\ref{task_def}. For the BERT-based retriever, we have 89.66\% recall for the top 3 retrieved facts and 93.63\% recall for the top 5. Using TF-IDF results in 82.91\% recall for the top 5 facts. We use the same retriever results for all retriever-generator based models. 
\begin{table}[t]
\small
\begin{center}
\resizebox{.48\textwidth}{!}{%
\begin{tabular}{lcc}
\toprule
\textbf{Baselines} & \textbf{Exe Acc} & \textbf{Prog Acc}\\
\midrule
TF-IDF + Single Op & 1.01 & 0.90 \\
\midrule
Retriever + Direct Generation & 0.30 & - \\
\midrule
Pre-Trained Longformer (base) & 21.90 & 20.48 \\
\midrule
Retriever + Seq2seq & 19.71 & 18.38 \\
\midrule
Retriever + NeRd (BERT-base) & 48.57 & 46.76 \\
\midrule
\midrule
FinQANet (FinBert) & 50.10 & 47.52 \\
\midrule
FinQANet (BERT-base) & 50.00 & 48.00 \\
\midrule
FinQANet (BERT-large) & 53.52 & 51.62 \\
\midrule
FinQANet (RoBERTa-base) & 56.10 & 54.38 \\
\midrule
FinQANet (RoBERTa-large) & \textbf{61.24} & \textbf{58.86} \\
\midrule
\midrule
FinQANet-Gold (RoBERTa-large) & 70.00 & 68.76 \\
\midrule
\midrule
Human Expert Performance & 91.16 & 87.49 \\
\midrule
General Crowd Performance & 50.68 & 48.17 \\
\bottomrule
\end{tabular}
}
\caption{The execution accuracy (Exe Acc) and program accuracy (Prog Acc) for all the models. Although our best system (61.24\%) outperforms the non-expert crowd (50.68\%), the significant accuracy gap between the model and human experts (91.16\%) motivates the need for future research.}
\vspace{-4mm}
\label{table:main_res}
\end{center}
\end{table}
\begin{table}[t]
\small
\begin{center}
\resizebox{.48\textwidth}{!}{
\begin{tabular}{lcc}
\toprule
\textbf{Methods} & \textbf{Exe Acc} & \textbf{Prog Acc}\\
\midrule
\textbf{full results} & \textbf{61.24} & \textbf{58.86} \\
\midrule
\multicolumn{2}{l}{\textbf{Necessity of table and text}} & \\
\midrule
table-only inference & 45.81 & 43.62 \\
\midrule
text-only inference & 15.80 & 15.33 \\
\midrule
\multicolumn{2}{l}{\textbf{Performances on table and text}} & \\
\midrule
table-only questions & 67.38 & 64.48 \\
\midrule
text-only questions & 54.86 & 53.70 \\
\midrule
table-text questions & 43.80 & 41.61 \\
\midrule
\multicolumn{2}{l}{\textbf{Performances regarding program steps}} & \\
\midrule
1 step programs & 67.61 & 65.28 \\
\midrule
2 step programs & 59.08 & 56.37 \\
\midrule
>2 step programs & 22.78 & 21.52 \\
\midrule
\textbf{Programs with constants} & 43.88 & 39.80 \\
\bottomrule
\end{tabular}
}
\caption{Performance breakdown of FinQANet (RoBERTa-large). The model benefits from using both table and text, as inferences on individual source yield much lower performance. FinQANet is better at answering table-only questions, and the questions that require more steps to solve are indeed more challenging to the model.}
\label{table:detail_res}
\end{center}
\end{table}
Directly generating the execution results gives near-zero scores, which indicates the necessity of generating the reasoning programs.
If without using the retriever-generator pipeline, but directly applying an end-to-end pre-trained Longformer model, the performance falls far behind.
Because longer inputs have more numbers which put more confusions on the program generator and thus make it harder to learn. Generally, the program generators using pre-trained models perform much better than the Seq2seq baseline, as there is language modeling knowledge that can also be used for the finance domain. And larger pre-trained models give better performance, as they tend to see more financial corpus during their pre-training. FinBert~\cite{DBLP:journals/corr/abs-1908-10063} is a pre-trained model for the finance domain; its main downstream tasks are sentiment analysis. The performance of using FinBert is no better than BERT-large, mostly because its pre-training corpus is limited (\textasciitilde30M words from news articles). 
Comparing FinQANet with the retriever + NeRd baseline~\cite{DBLP:conf/iclr/ChenLYZSL20}, it shows the improvements from learning the logical structure of the programs. We also run the program generator using the gold retriever result, shown as FinQANet-Gold. 
Another interesting observation is the comparisons with human performances. While there is still a large gap from the human expert upper bound, the best performing model already surpasses the general crowd performance. 

\subsection{Performance Breakdown}
\begin{figure*}[ht]
\centering
\includegraphics[width=\textwidth]{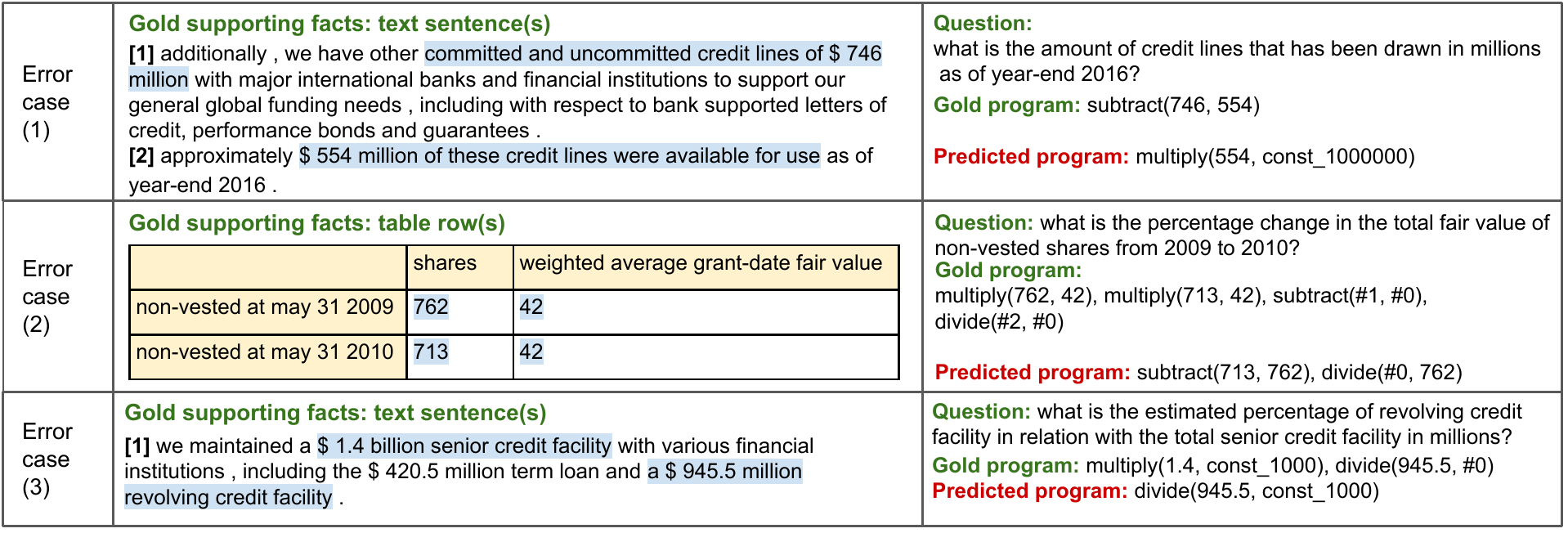}
\caption{Error cases. In these examples, the retriever results all correctly cover the gold facts; thus we only present the gold facts, gold program, and the predicted program to study the errors of the program generator. We give more error cases in Appendix C, including the cases for the retriever errors. \textbf{Example 1}: The financial knowledge to calculate the `credit lines that has been drawn'. \textbf{Example 2}: Complex reasoning of 4 steps. \textbf{Example 3}: Number unit conversion between `billion' and `million'.} 
\label{fig:case_study}
\end{figure*}

We conduct a set of performance breakdowns using the FinQANet (RoBERTa-large) model.
Table~\ref{table:detail_res} shows all the results.

\paragraph{Necessity of using both table and text.}
We run inferences taking facts only from a single source from the retriever. 
Inferences on individual source (table-only: 45.81\%, text-only: 15.80\%) are both far behind the full results (61.24\%). 

\paragraph{The model performs the best on the table-only questions.}
The model performs the best on table-only questions (67.38\%).
Tables tend to have more unified structures and might be easier for the model to learn.
Table~\ref{table:detail_res} also shows that the questions involving both tables and texts are the most challenging ones for the model (43.80\%).

\paragraph{Questions that need more than two steps to answer are challenging.}
The model has a low accuracy (22.78\%) on the questions that need three or more steps.
Meanwhile, not surprisingly, the questions that require only one step are the easiest.


\paragraph{Constants in programs.}
Many programs in \dataset contain constants as arguments.
A constant is often used to convert an English number word to another.
For example, we need first to use the constant ``1,000'' to convert ``1.5 billion'' to ``1,500 million'' so that it can be added with ``50 million''.
A constant is also used to explicate the implicit numbers hidden in the language.
For example, to calculate ``the average for the year 2012, 2013, and 2014'', the program needs to use the constant ``3''  as the denominator, which is not mentioned explicitly in the text.
As shown in Table~\ref{table:detail_res}, the programs with constants yield great challenges for our model, as the performance (43.88\%) is much lower than that of the whole set (61.24\%).

\subsection{Error Analysis}
We sample 50 error cases from the results of the FinQANet (RoBERTa-large) model and analyze them manually.
15\% of the errors are caused by the retriever, {\em e.g.}, missing facts.
Half of the rest are due to the lack of financial knowledge, such as the meaning of some terminology.
And the rest half are primarily numerical reasoning errors, 
including complex programs with multiple steps,
numerical unit conversions, or 
resolving the ordering and matching of the numbers and the years.
Many error cases
involve both the numerical reasoning problems and misunderstandings of financial knowledge.
We show three representative error cases in Figure~\ref{fig:case_study}.

\section{Conclusion and Future Work}

This paper introduces \dataset, a new expert-annotated QA dataset that aims to tackle numerical reasoning over real-world financial data.
The questions in \dataset pose great challenge for existing models to resolve domain-specific knowledge, as well as to acquire complex numerical reasoning abilities. 
We propose baseline frameworks and conduct comprehensive experiments and analysis. 
The results show that current large pre-trained models still fall far behind the human expert performance. This encourages potential future work on developing pre-training tasks for such realistic, complex application domains. We believe \dataset should serve as a valuable resource for the research community.

\section{Ethical Considerations}

\paragraph{Data Access and Licensing.}
We develop \textsc{FinQA} based on the publicly available earnings reports of S\&P 500 companies from 1999 to 2019, collected in the FinTabNet dataset~\cite{fintabnet}.
The FinTabNet dataset is publicly available under the CDLA-Permissive\footnote{CDLA-Permissive: https://cdla.dev/sharing-1-0/} license, which permits us to create additional annotations on top of the data (``Enhanced Data'', \S1.5 of CDLA) and publish the annotations (``Publish'', \S1.9 of CDLA).

\paragraph{Dataset Collection Process and Conditions.}
For the annotation of our \textsc{FinQA} dataset on Upwork, we first launch interviews of the task introduction with 4 example questions, which is paid as \$30, for them to try a few examples to get informed and familiar with the task. Then based on their consents to continue working on the large-scale job, we discuss with the workers to reach agreements on the compensation before starting the large-scale job. We pay around \$2.0 per question, and the hourly rates are discussed and agreed upon with both sides based on the working speed of different workers. Among all eleven US-based hires, the average hourly rate is \$35.0, and the minimum and maximum hourly rates are \$20 and \$50, respectively. The evaluation tasks follow the similar procedure, and each question is paid as \$2.0. 

\paragraph{IRB (Institutional Review Board) Approval.}
This project is approved by our Institutional Review Board (IRB). The systems trained using our dataset are primarily intended to be used as augmenting human decision-making in financial analysis, but not as a replacement of human experts.

\section*{Acknowledgment}
We thank the anonymous reviewers for their thoughtful comments. This research was supported by the J.P. Morgan Faculty research award. The authors are solely responsible for the contents of the paper and the opinions expressed in this publication do not reflect those of the funding agencies.

\bibliography{emnlp2021}

\begin{thebibliography}{32}
\expandafter\ifx\csname natexlab\endcsname\relax\def\natexlab#1{#1}\fi

\bibitem[{Akhtar et~al.(2017)Akhtar, Kumar, Ghosal, Ekbal, and
  Bhattacharyya}]{DBLP:conf/emnlp/AkhtarKGEB17}
Md.~Shad Akhtar, Abhishek Kumar, Deepanway Ghosal, Asif Ekbal, and Pushpak
  Bhattacharyya. 2017.
\newblock \href {https://doi.org/10.18653/v1/d17-1057} {A multilayer perceptron
  based ensemble technique for fine-grained financial sentiment analysis}.
\newblock In \emph{Proceedings of the 2017 Conference on Empirical Methods in
  Natural Language Processing, {EMNLP} 2017, Copenhagen, Denmark, September
  9-11, 2017}, pages 540--546. Association for Computational Linguistics.

\bibitem[{Alberti et~al.(2019)Alberti, Lee, and
  Collins}]{DBLP:journals/corr/abs-1901-08634}
Chris Alberti, Kenton Lee, and Michael Collins. 2019.
\newblock \href {http://arxiv.org/abs/1901.08634} {A {BERT} baseline for the
  natural questions}.
\newblock \emph{CoRR}, abs/1901.08634.

\bibitem[{Amini et~al.(2019)Amini, Gabriel, Lin, Koncel{-}Kedziorski, Choi, and
  Hajishirzi}]{DBLP:conf/naacl/AminiGLKCH19}
Aida Amini, Saadia Gabriel, Shanchuan Lin, Rik Koncel{-}Kedziorski, Yejin Choi,
  and Hannaneh Hajishirzi. 2019.
\newblock \href {https://doi.org/10.18653/v1/n19-1245} {Mathqa: Towards
  interpretable math word problem solving with operation-based formalisms}.
\newblock In \emph{Proceedings of the 2019 Conference of the North American
  Chapter of the Association for Computational Linguistics: Human Language
  Technologies, {NAACL-HLT} 2019, Minneapolis, MN, USA, June 2-7, 2019, Volume
  1 (Long and Short Papers)}, pages 2357--2367. Association for Computational
  Linguistics.

\bibitem[{Araci(2019)}]{DBLP:journals/corr/abs-1908-10063}
Dogu Araci. 2019.
\newblock \href {http://arxiv.org/abs/1908.10063} {Finbert: Financial sentiment
  analysis with pre-trained language models}.
\newblock \emph{CoRR}, abs/1908.10063.

\bibitem[{Beltagy et~al.(2020)Beltagy, Peters, and
  Cohan}]{DBLP:journals/corr/abs-2004-05150}
Iz~Beltagy, Matthew~E. Peters, and Arman Cohan. 2020.
\newblock \href {http://arxiv.org/abs/2004.05150} {Longformer: The
  long-document transformer}.
\newblock \emph{CoRR}, abs/2004.05150.

\bibitem[{Chen et~al.(2020{\natexlab{a}})Chen, Huang, Palangi, Smolensky,
  Forbus, and Gao}]{DBLP:conf/icml/ChenHPSFG20}
Kezhen Chen, Qiuyuan Huang, Hamid Palangi, Paul Smolensky, Kenneth~D. Forbus,
  and Jianfeng Gao. 2020{\natexlab{a}}.
\newblock \href {http://proceedings.mlr.press/v119/chen20g.html} {Mapping
  natural-language problems to formal-language solutions using structured
  neural representations}.
\newblock In \emph{Proceedings of the 37th International Conference on Machine
  Learning, {ICML} 2020, 13-18 July 2020, Virtual Event}, volume 119 of
  \emph{Proceedings of Machine Learning Research}, pages 1566--1575. {PMLR}.

\bibitem[{Chen et~al.(2020{\natexlab{b}})Chen, Wang, Chen, Zhang, Wang, Li,
  Zhou, and Wang}]{DBLP:conf/iclr/ChenWCZWLZW20}
Wenhu Chen, Hongmin Wang, Jianshu Chen, Yunkai Zhang, Hong Wang, Shiyang Li,
  Xiyou Zhou, and William~Yang Wang. 2020{\natexlab{b}}.
\newblock \href {https://openreview.net/forum?id=rkeJRhNYDH} {Tabfact: {A}
  large-scale dataset for table-based fact verification}.
\newblock In \emph{8th International Conference on Learning Representations,
  {ICLR} 2020, Addis Ababa, Ethiopia, April 26-30, 2020}. OpenReview.net.

\bibitem[{Chen et~al.(2020{\natexlab{c}})Chen, Zha, Chen, Xiong, Wang, and
  Wang}]{DBLP:conf/emnlp/ChenZCXWW20}
Wenhu Chen, Hanwen Zha, Zhiyu Chen, Wenhan Xiong, Hong Wang, and William~Yang
  Wang. 2020{\natexlab{c}}.
\newblock \href {https://doi.org/10.18653/v1/2020.findings-emnlp.91} {Hybridqa:
  {A} dataset of multi-hop question answering over tabular and textual data}.
\newblock In \emph{Proceedings of the 2020 Conference on Empirical Methods in
  Natural Language Processing: Findings, {EMNLP} 2020, Online Event, 16-20
  November 2020}, pages 1026--1036. Association for Computational Linguistics.

\bibitem[{Chen et~al.(2020{\natexlab{d}})Chen, Liang, Yu, Zhou, Song, and
  Le}]{DBLP:conf/iclr/ChenLYZSL20}
Xinyun Chen, Chen Liang, Adams~Wei Yu, Denny Zhou, Dawn Song, and Quoc~V. Le.
  2020{\natexlab{d}}.
\newblock \href {https://openreview.net/forum?id=ryxjnREFwH} {Neural symbolic
  reader: Scalable integration of distributed and symbolic representations for
  reading comprehension}.
\newblock In \emph{8th International Conference on Learning Representations,
  {ICLR} 2020, Addis Ababa, Ethiopia, April 26-30, 2020}. OpenReview.net.

\bibitem[{Day and Lee(2016)}]{DBLP:conf/asunam/DayL16}
Min{-}Yuh Day and Chia{-}Chou Lee. 2016.
\newblock \href {https://doi.org/10.1109/ASONAM.2016.7752381} {Deep learning
  for financial sentiment analysis on finance news providers}.
\newblock In \emph{2016 {IEEE/ACM} International Conference on Advances in
  Social Networks Analysis and Mining, {ASONAM} 2016, San Francisco, CA, USA,
  August 18-21, 2016}, pages 1127--1134. {IEEE} Computer Society.

\bibitem[{Devlin et~al.(2019)Devlin, Chang, Lee, and
  Toutanova}]{DBLP:conf/naacl/DevlinCLT19}
Jacob Devlin, Ming{-}Wei Chang, Kenton Lee, and Kristina Toutanova. 2019.
\newblock \href {https://doi.org/10.18653/v1/n19-1423} {{BERT:} pre-training of
  deep bidirectional transformers for language understanding}.
\newblock In \emph{Proceedings of the 2019 Conference of the North American
  Chapter of the Association for Computational Linguistics: Human Language
  Technologies, {NAACL-HLT} 2019, Minneapolis, MN, USA, June 2-7, 2019, Volume
  1 (Long and Short Papers)}, pages 4171--4186. Association for Computational
  Linguistics.

\bibitem[{Dua et~al.(2019)Dua, Wang, Dasigi, Stanovsky, Singh, and
  Gardner}]{DBLP:conf/naacl/DuaWDSS019}
Dheeru Dua, Yizhong Wang, Pradeep Dasigi, Gabriel Stanovsky, Sameer Singh, and
  Matt Gardner. 2019.
\newblock \href {https://doi.org/10.18653/v1/n19-1246} {{DROP:} {A} reading
  comprehension benchmark requiring discrete reasoning over paragraphs}.
\newblock In \emph{Proceedings of the 2019 Conference of the North American
  Chapter of the Association for Computational Linguistics: Human Language
  Technologies, {NAACL-HLT} 2019, Minneapolis, MN, USA, June 2-7, 2019, Volume
  1 (Long and Short Papers)}, pages 2368--2378. Association for Computational
  Linguistics.

\bibitem[{Han et~al.(2018)Han, Barman, Hayes, Du, Burgin, and
  Wan}]{DBLP:conf/acl/HanBHDBW18}
Jingguang Han, Utsab Barman, Jer Hayes, Jinhua Du, Edward Burgin, and Dadong
  Wan. 2018.
\newblock \href {https://doi.org/10.18653/v1/P18-4007} {Nextgen {AML:}
  distributed deep learning based language technologies to augment anti money
  laundering investigation}.
\newblock In \emph{Proceedings of {ACL} 2018, Melbourne, Australia, July 15-20,
  2018, System Demonstrations}, pages 37--42. Association for Computational
  Linguistics.

\bibitem[{Jerven(2013)}]{jerven2013poor}
Morten Jerven. 2013.
\newblock \emph{Poor numbers: how we are misled by African development
  statistics and what to do about it}.
\newblock Cornell University Press.

\bibitem[{Kim et~al.(2020)Kim, Ki, Lee, and Gweon}]{DBLP:conf/emnlp/KimKLG20}
Bugeun Kim, Kyung~Seo Ki, Donggeon Lee, and Gahgene Gweon. 2020.
\newblock \href {https://doi.org/10.18653/v1/2020.emnlp-main.308} {Point to the
  expression: Solving algebraic word problems using the expression-pointer
  transformer model}.
\newblock In \emph{Proceedings of the 2020 Conference on Empirical Methods in
  Natural Language Processing, {EMNLP} 2020, Online, November 16-20, 2020},
  pages 3768--3779. Association for Computational Linguistics.

\bibitem[{Kingma and Ba(2015)}]{DBLP:journals/corr/KingmaB14}
Diederik~P. Kingma and Jimmy Ba. 2015.
\newblock \href {http://arxiv.org/abs/1412.6980} {Adam: {A} method for
  stochastic optimization}.
\newblock In \emph{3rd International Conference on Learning Representations,
  {ICLR} 2015, San Diego, CA, USA, May 7-9, 2015, Conference Track
  Proceedings}.

\bibitem[{Koncel{-}Kedziorski et~al.(2016)Koncel{-}Kedziorski, Roy, Amini,
  Kushman, and Hajishirzi}]{DBLP:conf/naacl/Koncel-Kedziorski16}
Rik Koncel{-}Kedziorski, Subhro Roy, Aida Amini, Nate Kushman, and Hannaneh
  Hajishirzi. 2016.
\newblock \href {https://doi.org/10.18653/v1/n16-1136} {{MAWPS:} {A} math word
  problem repository}.
\newblock In \emph{{NAACL} {HLT} 2016, The 2016 Conference of the North
  American Chapter of the Association for Computational Linguistics: Human
  Language Technologies, San Diego California, USA, June 12-17, 2016}, pages
  1152--1157. The Association for Computational Linguistics.

\bibitem[{Lange et~al.(2016)Lange, Lenglet, and Seyfert}]{lange2016cultures}
Ann-Christina Lange, Marc Lenglet, and Robert Seyfert. 2016.
\newblock Cultures of high-frequency trading: Mapping the landscape of
  algorithmic developments in contemporary financial markets.
\newblock \emph{Economy and Society}, 45(2):149--165.

\bibitem[{Liu et~al.(2019)Liu, Ott, Goyal, Du, Joshi, Chen, Levy, Lewis,
  Zettlemoyer, and Stoyanov}]{DBLP:journals/corr/abs-1907-11692}
Yinhan Liu, Myle Ott, Naman Goyal, Jingfei Du, Mandar Joshi, Danqi Chen, Omer
  Levy, Mike Lewis, Luke Zettlemoyer, and Veselin Stoyanov. 2019.
\newblock \href {http://arxiv.org/abs/1907.11692} {Roberta: {A} robustly
  optimized {BERT} pretraining approach}.
\newblock \emph{CoRR}, abs/1907.11692.

\bibitem[{Liu et~al.(2020)Liu, Huang, Huang, Li, and
  Zhao}]{DBLP:conf/ijcai/0001HH0Z20}
Zhuang Liu, Degen Huang, Kaiyu Huang, Zhuang Li, and Jun Zhao. 2020.
\newblock \href {https://doi.org/10.24963/ijcai.2020/622} {Finbert: {A}
  pre-trained financial language representation model for financial text
  mining}.
\newblock In \emph{Proceedings of the Twenty-Ninth International Joint
  Conference on Artificial Intelligence, {IJCAI} 2020}, pages 4513--4519.
  ijcai.org.

\bibitem[{MacKenzie(2008)}]{mackenzie2008engine}
Donald MacKenzie. 2008.
\newblock \emph{An engine, not a camera: How financial models shape markets}.
\newblock Mit Press.

\bibitem[{MacKenzie et~al.(2012)MacKenzie, Beunza, Millo, and
  Pardo-Guerra}]{mackenzie2012drilling}
Donald MacKenzie, Daniel Beunza, Yuval Millo, and Juan~Pablo Pardo-Guerra.
  2012.
\newblock Drilling through the allegheny mountains: Liquidity, materiality and
  high-frequency trading.
\newblock \emph{Journal of cultural economy}, 5(3):279--296.

\bibitem[{Nourbakhsh and Bang(2019)}]{DBLP:journals/corr/abs-1908-09156}
Armineh Nourbakhsh and Grace Bang. 2019.
\newblock \href {http://arxiv.org/abs/1908.09156} {A framework for anomaly
  detection using language modeling, and its applications to finance}.
\newblock \emph{CoRR}, abs/1908.09156.

\bibitem[{Pasupat and Liang(2015)}]{DBLP:conf/acl/PasupatL15}
Panupong Pasupat and Percy Liang. 2015.
\newblock \href {https://doi.org/10.3115/v1/p15-1142} {Compositional semantic
  parsing on semi-structured tables}.
\newblock In \emph{Proceedings of the 53rd Annual Meeting of the Association
  for Computational Linguistics and the 7th International Joint Conference on
  Natural Language Processing of the Asian Federation of Natural Language
  Processing, {ACL} 2015, July 26-31, 2015, Beijing, China, Volume 1: Long
  Papers}, pages 1470--1480. The Association for Computer Linguistics.

\bibitem[{Rajpurkar et~al.(2018)Rajpurkar, Jia, and
  Liang}]{DBLP:conf/acl/RajpurkarJL18}
Pranav Rajpurkar, Robin Jia, and Percy Liang. 2018.
\newblock \href {https://doi.org/10.18653/v1/P18-2124} {Know what you don't
  know: Unanswerable questions for squad}.
\newblock In \emph{Proceedings of the 56th Annual Meeting of the Association
  for Computational Linguistics, {ACL} 2018, Melbourne, Australia, July 15-20,
  2018, Volume 2: Short Papers}, pages 784--789. Association for Computational
  Linguistics.

\bibitem[{Talmor and Berant(2018)}]{DBLP:conf/naacl/TalmorB18}
Alon Talmor and Jonathan Berant. 2018.
\newblock \href {https://doi.org/10.18653/v1/n18-1059} {The web as a
  knowledge-base for answering complex questions}.
\newblock In \emph{Proceedings of the 2018 Conference of the North American
  Chapter of the Association for Computational Linguistics: Human Language
  Technologies, {NAACL-HLT} 2018, New Orleans, Louisiana, USA, June 1-6, 2018,
  Volume 1 (Long Papers)}, pages 641--651. Association for Computational
  Linguistics.

\bibitem[{Wang et~al.(2013)Wang, Tsai, Liu, and
  Chang}]{DBLP:conf/ijcnlp/WangTLC13}
Chuan{-}Ju Wang, Ming{-}Feng Tsai, Tse Liu, and Chin{-}Ting Chang. 2013.
\newblock \href {https://www.aclweb.org/anthology/I13-1097/} {Financial
  sentiment analysis for risk prediction}.
\newblock In \emph{Sixth International Joint Conference on Natural Language
  Processing, {IJCNLP} 2013, Nagoya, Japan, October 14-18, 2013}, pages
  802--808. Asian Federation of Natural Language Processing / {ACL}.

\bibitem[{Wang et~al.(2019)Wang, Zhang, Li, Zong, and
  Li}]{DBLP:conf/emnlp/WangZLZL19}
Weikang Wang, Jiajun Zhang, Qian Li, Chengqing Zong, and Zhifei Li. 2019.
\newblock \href {https://doi.org/10.18653/v1/D19-1185} {Are you for real?
  detecting identity fraud via dialogue interactions}.
\newblock In \emph{Proceedings of the 2019 Conference on Empirical Methods in
  Natural Language Processing and the 9th International Joint Conference on
  Natural Language Processing, {EMNLP-IJCNLP} 2019, Hong Kong, China, November
  3-7, 2019}, pages 1762--1771. Association for Computational Linguistics.

\bibitem[{Yang et~al.(2020)Yang, Uy, and
  Huang}]{DBLP:journals/corr/abs-2006-08097}
Yi~Yang, Mark Christopher~Siy Uy, and Allen Huang. 2020.
\newblock \href {http://arxiv.org/abs/2006.08097} {Finbert: {A} pretrained
  language model for financial communications}.
\newblock \emph{CoRR}, abs/2006.08097.

\bibitem[{Yang et~al.(2018)Yang, Qi, Zhang, Bengio, Cohen, Salakhutdinov, and
  Manning}]{DBLP:conf/emnlp/Yang0ZBCSM18}
Zhilin Yang, Peng Qi, Saizheng Zhang, Yoshua Bengio, William~W. Cohen, Ruslan
  Salakhutdinov, and Christopher~D. Manning. 2018.
\newblock \href {https://doi.org/10.18653/v1/d18-1259} {Hotpotqa: {A} dataset
  for diverse, explainable multi-hop question answering}.
\newblock In \emph{Proceedings of the 2018 Conference on Empirical Methods in
  Natural Language Processing, Brussels, Belgium, October 31 - November 4,
  2018}, pages 2369--2380. Association for Computational Linguistics.

\bibitem[{Yu et~al.(2018)Yu, Zhang, Yang, Yasunaga, Wang, Li, Ma, Li, Yao,
  Roman, Zhang, and Radev}]{DBLP:conf/emnlp/YuZYYWLMLYRZR18}
Tao Yu, Rui Zhang, Kai Yang, Michihiro Yasunaga, Dongxu Wang, Zifan Li, James
  Ma, Irene Li, Qingning Yao, Shanelle Roman, Zilin Zhang, and Dragomir~R.
  Radev. 2018.
\newblock \href {https://doi.org/10.18653/v1/d18-1425} {Spider: {A} large-scale
  human-labeled dataset for complex and cross-domain semantic parsing and
  text-to-sql task}.
\newblock In \emph{Proceedings of the 2018 Conference on Empirical Methods in
  Natural Language Processing, Brussels, Belgium, October 31 - November 4,
  2018}, pages 3911--3921. Association for Computational Linguistics.

\bibitem[{Zheng et~al.(2021)Zheng, Burdick, Popa, Zhong, and Wang}]{fintabnet}
Xinyi Zheng, Doug Burdick, Lucian Popa, Peter Zhong, and Nancy Xin~Ru Wang.
  2021.
\newblock Global table extractor (gte): A framework for joint table
  identification and cell structure recognition using visual context.
\newblock \emph{Winter Conference for Applications in Computer Vision (WACV)}.

\end{thebibliography}
\bibliographystyle{acl_natbib}

\section*{Appendix A: Operation Definitions}

We describe all the operations in Table~\ref{table:def_op}.

\begin{table*}[ht]
\resizebox{\textwidth}{!}{%
\begin{tabular}{l|l|l|l}
\toprule
Name & Arguments & Output & Description \\
\midrule
add & number1, number2 & number & add two numbers: $number1 + number2$ \\
\midrule
subtract & number1, number2 & number & subtract two numbers: $number1 - number2$ \\
\midrule
multiply & number1, number2 & number & multiply two numbers: $number1 \cdot number2$ \\
\midrule
divide & number1, number2 & number & multiply two numbers: $number1 / number2$ \\
\midrule
exp & number1, number2 & number & exponential: $number1^{number2}$ \\
\midrule
greater & number1, number2 & bool & comparison: $number1 > number2$ \\
\midrule
table-sum & table header & number & the summation of one table row \\
\midrule
table-average & table header & number & the average of one table row \\
\midrule
table-max & table header & number & the maximum number of one table row \\
\midrule
table-min & table header & number & the minimum number of one table row \\
\bottomrule
\end{tabular}
}
\caption{Definitions of all operations}
\label{table:def_op}
\end{table*}

\section*{Appendix B: Experiment Details}

All the validation results of the baselines are shown in Table~\ref{table:main_res_valid}. The trainings of all models are conducted on TITAN RTX GPUs. All the implementation and pre-trained models are based on the huggingface transformers library. We use the Adam optimizer~\cite{DBLP:journals/corr/KingmaB14}. The parameter settings are the following:

\noindent \textbf{Retriever}
The learning rate is set as 3e-5, with batch size of 16. 

\noindent \textbf{TF-IDF + Single Op}
We use the TF-IDF from the Scikit-learn library. 

\noindent \textbf{FinQANet}
The learning rate is set as 1e-5. For Bert-base, Roberta-base, and finBert we use batch size of 32; For Bert-large and RoBerta-large we use batch size of 16 due to GPU memory constraints. 

\noindent \textbf{Retriever + Seq2seq}
A bidirectional LSTM is used for encoding the input, then an LSTM is used for decoding with attention. Learning rate is set as 1e-3, hidden size as 100. 

\noindent \textbf{Retriever + NeRd}
The parameter settings are the same as FinQANet. 

\noindent \textbf{Pre-Trained Longformer}
We truncate the maximum input length as 2,000. The learning rate is set as 2e-5, with batch size of 16 due to GPU memory constraints. 

For more modeling details refer to our released code. 

\begin{table}[htbp]
\small
\begin{center}
\resizebox{.48\textwidth}{!}{%
\begin{tabular}{lcc}
\toprule
\textbf{Baselines} & \makecell[l]{\textbf{Execution} \\\textbf{Accuracy (\%)}} & \makecell[l]{\textbf{Program} \\\textbf{Accuracy (\%)}}\\
\midrule
TF-IDF + Single Op & 1.65 & 1.65 \\
\midrule
\makecell[l]{Retriever + \\Direct Generation} & 0.87 & - \\
\midrule
\makecell[l]{Pre-Trained \\Longformer (base)} & 23.83 & 22.56 \\
\midrule
Retriever + Seq2seq & 18.76 & 17.52 \\
\midrule
\makecell[l]{Retriever + \\NeRd (BERT-base)} & 47.53 & 45.37 \\
\midrule
\midrule
FinQANet (FinBert) & 46.64 & 44.11 \\
\midrule
FinQANet (BERT-base) & 49.91 & 47.15 \\
\midrule
FinQANet (BERT-large) & 53.86 & 50.95 \\
\midrule
FinQANet (RoBerta-base) & 56.27 & 53.49 \\
\midrule
FinQANet (RoBerta-large) & 61.22 & 58.05 \\
\bottomrule
\end{tabular}
}
\caption{Results on validation set
}
\label{table:main_res_valid}
\end{center}
\end{table}

\section*{Appendix C: Case Studies}
Here we provide more case studies with the full input reports. For all the examples the gold evidence is highlighted in blue.

\begin{figure*}[ht]
\centering
\includegraphics[width=\textwidth]{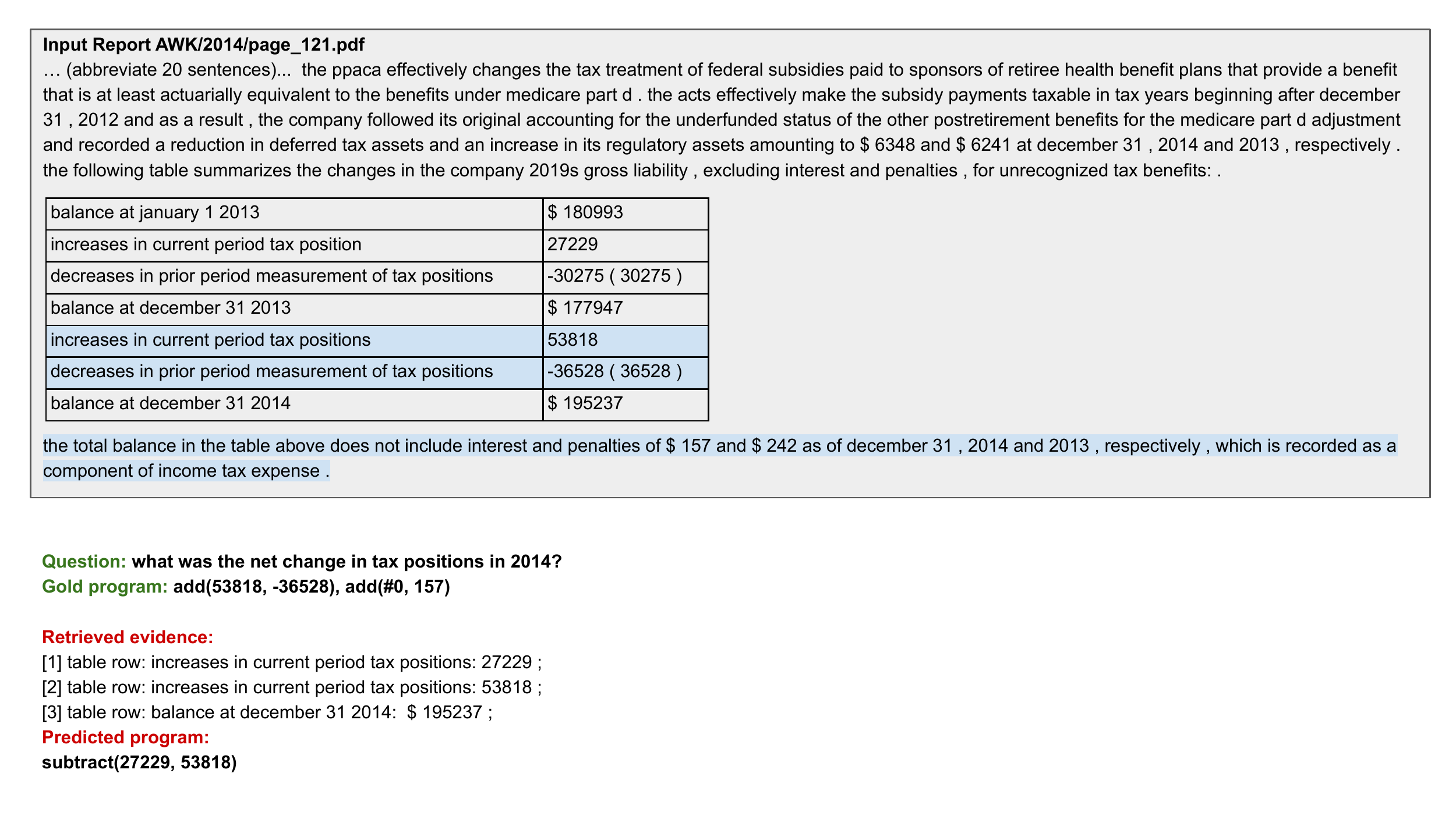}
\caption{Error case study 1: The net change in the tax position is the sum of the increase and the decrease plus the penalties and interest. The model lacks this finance knowledge, thus the retriever fails to retrieve the correct table rows and sentences. Another challenging point is the table understanding, since in this case, it's hard to distinguish the retrieved two table rows for the year 2013 or 2014, using our method that regards each table row as basic unit. The model needs to look at the full table to get this global information. } 
\label{fig:app_1}
\end{figure*}


\begin{figure*}[ht]
\centering
\includegraphics[width=\textwidth]{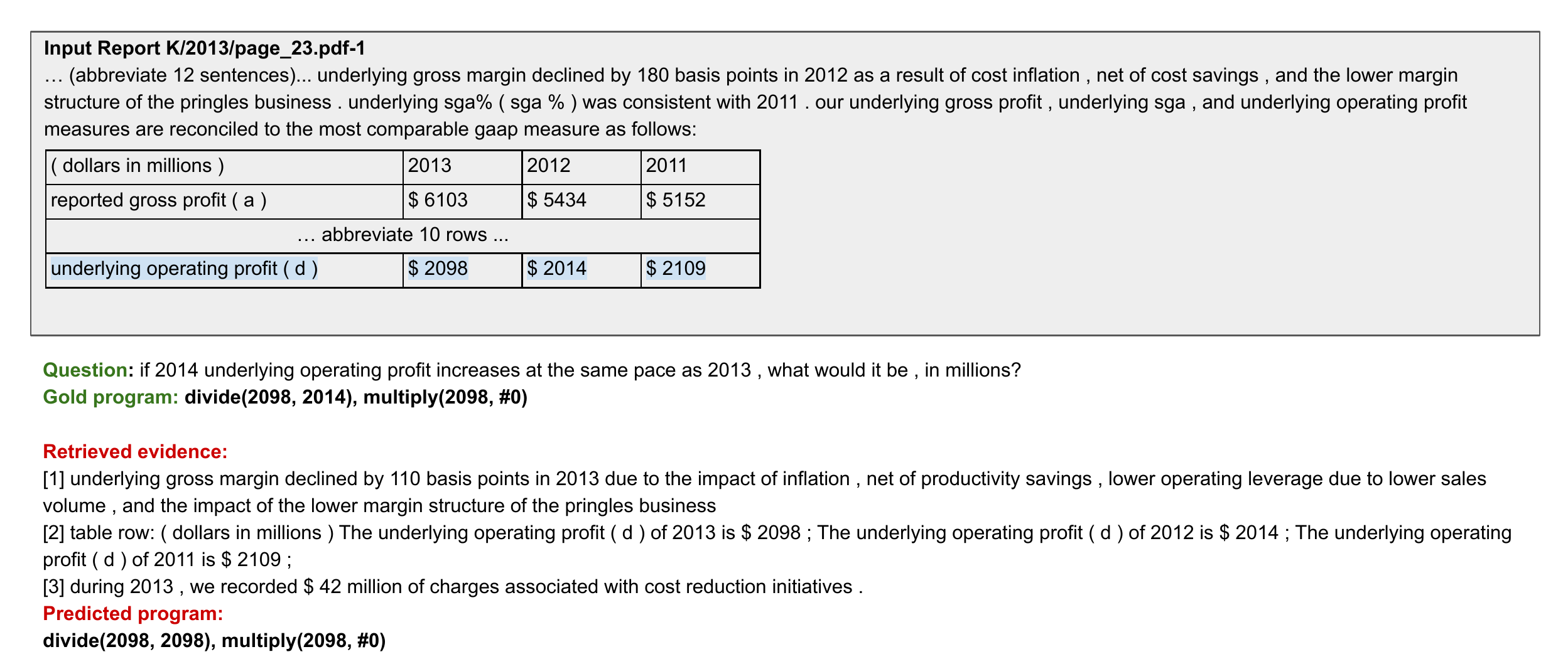}
\caption{Error case study 2: Complex numerical reasoning. } 
\label{fig:app_3}
\end{figure*}

\section*{Appendix D: Annotation Interface}
\begin{figure*}[ht]
\centering
\includegraphics[width=\textwidth]{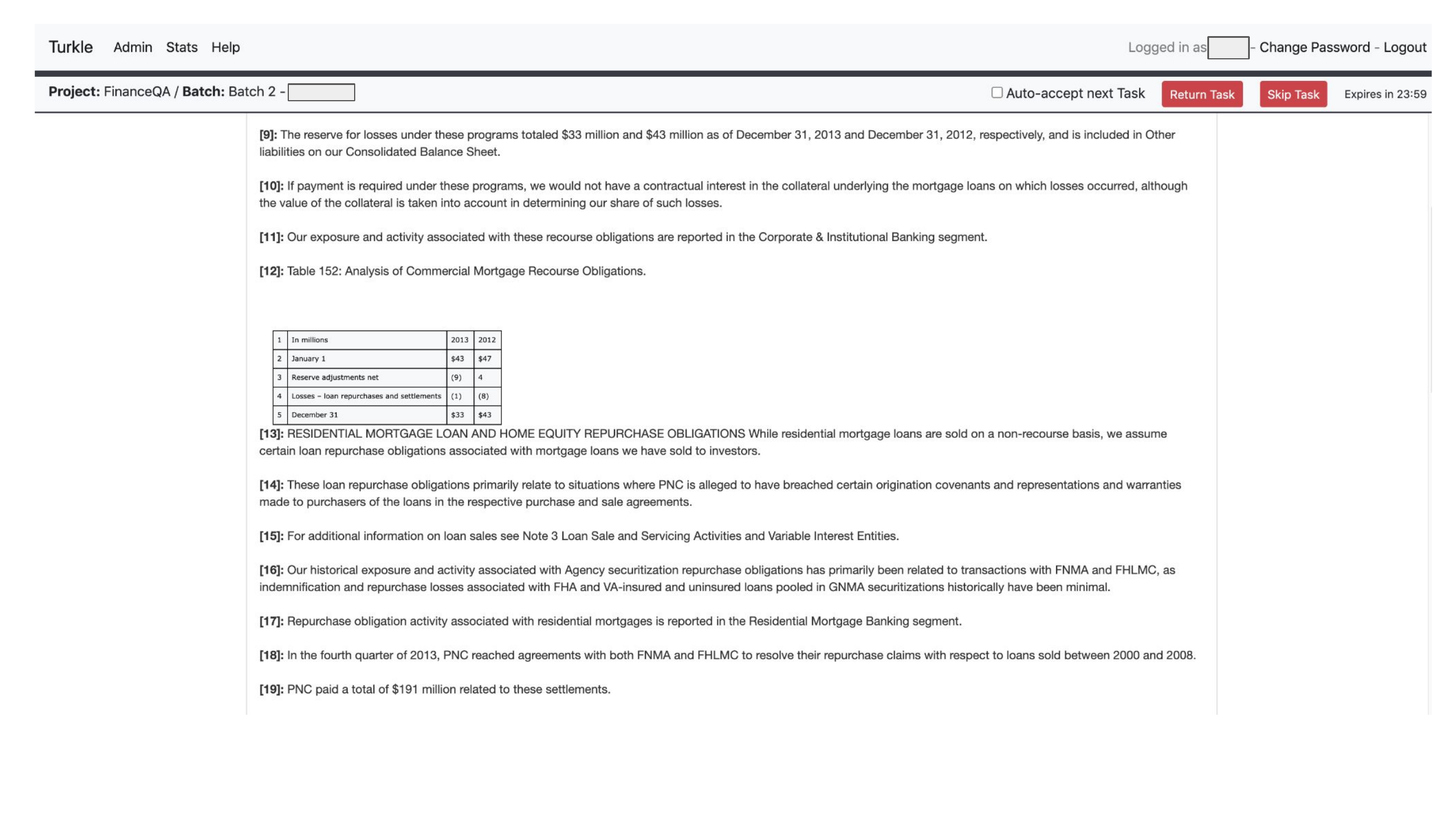}
\caption{Annotation interface: Display report. } 
\label{fig:ui_0}
\end{figure*}

\begin{figure*}[ht]
\centering
\includegraphics[width=\textwidth]{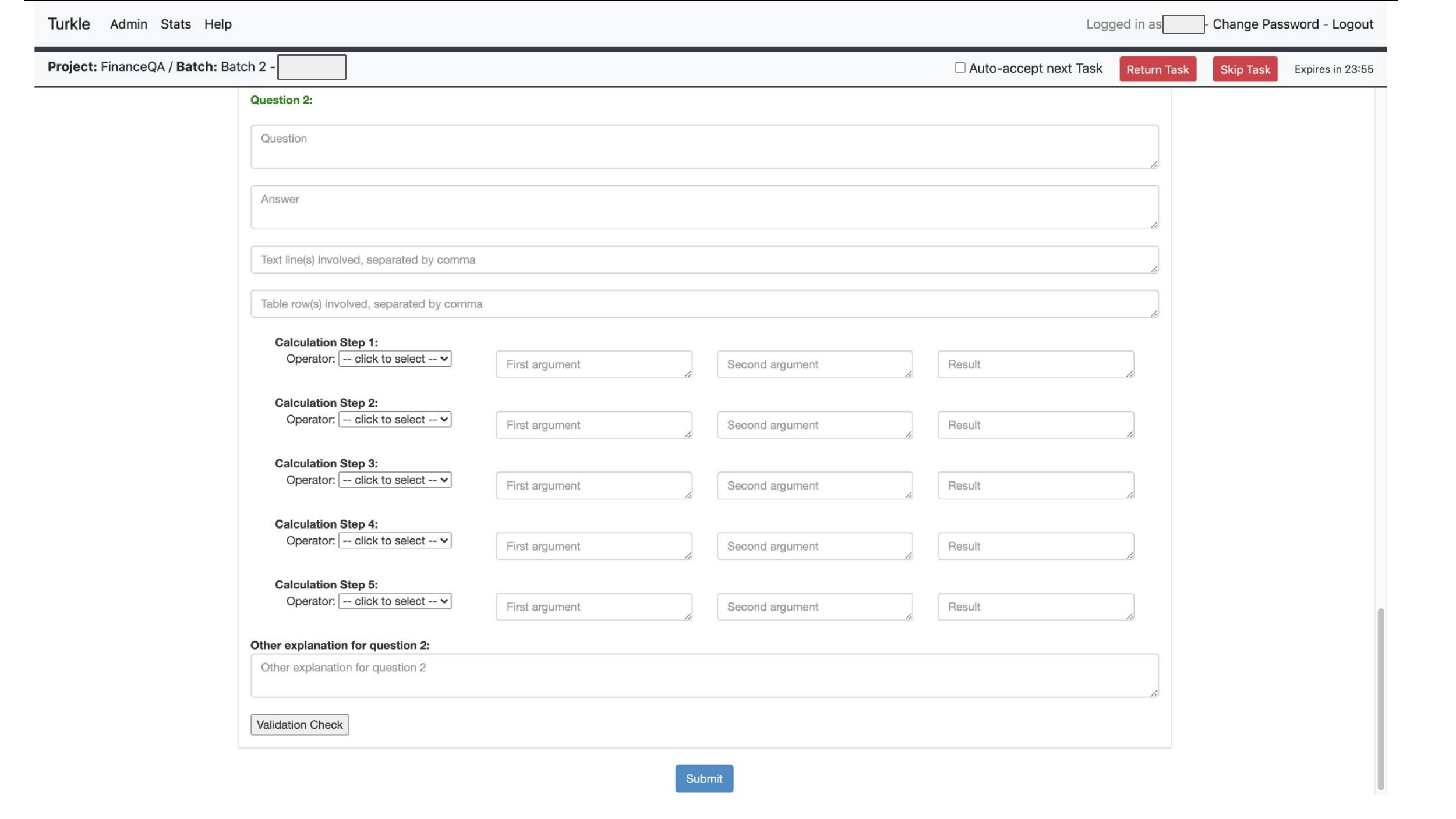}
\caption{Annotation interface: Annotator input fields. } 
\label{fig:ui_1}
\end{figure*}

We use Turkle\footnote{https://github.com/hltcoe/turkle} to build our annotation platform, which is a Django-based web application that can run in a local server. Figure~\ref{fig:ui_0} and Figure~\ref{fig:ui_1} show our annotation interface. After the annotators finish one example, they will use the validation check button to automatically check the validity of their inputs.


\end{document}